\documentclass{article}

\usepackage{arxiv}

\usepackage[utf8]{inputenc} 
\usepackage[T1]{fontenc}    
\usepackage{hyperref}       
\usepackage{url}            
\usepackage{booktabs}       
\usepackage{amsfonts}       
\usepackage{nicefrac}       
\usepackage{microtype}      
\usepackage{lipsum}		
\usepackage{graphicx}
\usepackage{natbib}
\usepackage{doi}

\usepackage{algorithm}
\usepackage{algpseudocode}
\usepackage{amsmath,amsfonts}
\usepackage{cleveref}
\usepackage{adjustbox}
\usepackage{multirow}
\usepackage[table]{xcolor}
\usepackage{caption}
\usepackage{subcaption}
\usepackage{booktabs}
\newcommand{\mathE}{\mathbb{E}}

\DeclareMathOperator*{\argmin}{arg\,min}

\usepackage{cleveref}

\newcommand{\eg}{\textit{e.g\@.}}
\newcommand{\etc}{\textit{etc\@.}}

\newcommand{\ie}{\textit{i.e\@.}}
\newcommand{\vs}{\textit{vs\@.}}

\newcommand{\btheta}{\boldsymbol{\theta}}

\title{Model Architecture Adaption for Bayesian Neural Networks}

\author{
	Duo Wang  \thanks{equal contribution} \\
	University of Cambridge \\
	wd263@cam.ac.uk \\
	\And
	Yiren Zhao \footnotemark[1] \\
	University of Cambridge \\
	yiren.zhao@cl.cam.ac.uk \\
	\And
	Ilia Shumailov \\
	University of Cambridge \\
	ilia.shumailov@cl.cam.ac.uk \\
	\And
	Robert Mullins \\
	University of Cambridge \\
	robert.mullins@cl.cam.ac.uk \\
}

\date{}

\renewcommand{\shorttitle}{Model Architecture Adaption for Bayesian Neural Networks}

\hypersetup{
pdftitle={A template for the arxiv style},
pdfsubject={q-bio.NC, q-bio.QM},
pdfauthor={David S.~Hippocampus, Elias D.~Striatum},
pdfkeywords={First keyword, Second keyword, More},
}

\begin{document}
\maketitle

\begin{abstract}
	Bayesian Neural Networks (BNNs) offer a mathematically grounded framework to quantify the uncertainty of model predictions but come with a prohibitive computation cost for both training and inference.
	In this work, we show a novel network architecture search (NAS) that optimizes BNNs for both accuracy and uncertainty while having a reduced inference latency. Different from canonical NAS that optimizes solely for in-distribution likelihood, the proposed scheme searches for the uncertainty performance using both in- and out-of-distribution data. Our method is able to search for the correct placement of Bayesian layer(s) in a network. 
	In our experiments, the searched models show comparable uncertainty quantification ability and accuracy compared to the state-of-the-art (deep ensemble). In addition, the searched models use only a fraction of the runtime compared to many popular BNN baselines, reducing the inference runtime cost by $2.98 \times$ and $2.92 \times$ respectively on the CIFAR10 dataset when compared to MCDropout and deep ensemble.

\end{abstract}


\section{Introduction}

Deep Neural Networks (DNNs) are prone to over-fitting and often tend to make overly confident predictions \cite{kristiadi2020being}, especially with inputs that are out of the original training data distribution.
Ideally, we would like to have a principled DNN that assigns low confidence scores to samples that cannot be well interpreted from the training information and high scores to inputs that are in the training manifold.
The ability for DNNs to say how certain they are in their predictions is particularly important for applications involving critical decision makings, such as healthcare and finance \cite{jiang2017artificial}. 

Bayesian Neural Networks (BNNs) offer a way to quantify uncertainty by approximating the posterior distribution over their model parameters.
The general framework is to construct or re-form a stochastic component in the network, \eg~stochastic weights \cite{blundell2015weight} or stochastic activations \cite{gal2016dropout}, to simulate the effect of having multiple models $\btheta$ with their associated probability distribution $q(\btheta)$.
When using a BNN for prediction, a set of possible models $\btheta_i$ is sampled and is used to produce a set of output values that can later be aggregated to provide an uncertainty metric  \cite{jospin2020hands}. Naturally, at test time, this Monte-Carlo approach requires multiple BNN inference runs for a single input data point, causing a huge inference runtime overhead when considering deploying them into real production systems.

A practical approach to alleviating the runtime overhead is to adjust the BNN network architecture. For instance, it is popular to apply Bayesian inference on the ($n$-)last layer(s) only, this is equivalent to having a point estimate network followed by a shallow BNN \cite{jospin2020hands}. Moreover, the architecture modifications can include other design dimensions in the network architecture design space, including channel lengths, kernel sizes, \etc~
This prompts the following question: \textit{How can we automatically optimize BNN model architectures for both accuracy and uncertainty measurements?}

In this work, we demonstrate a novel network architecture search (NAS) algorithm that optimizes not only the in-distribution (i.d) data accuracy but also uncertainty measurements for both i.d and out-of-distribution (o.o.d) data.
In particular, this work has the following contributions:

\begin{itemize}
	\item We propose a novel network architecture search framework focusing on both accuracy and uncertainty measurements. Classic NAS optimizes solely for in-distribution likelihood, on the contrary, the proposed optimization framework searches for suitable architectures with the best uncertainty performance using both i.d and o.o.d data. We also propose a principled way of generating o.o.d data for NAS using Variational Auto-Encoders. 
	\item We define a new search space for BNNs. The searched network can have a subset of layers ($n$-last layers) being stochastic while having the rest of the network being deterministic. We demonstrate how this design helps networks to achieve a better runtime when performing Bayesian inference. 
	\item We empirically evaluate our method on multiple datasets and a wide collection of out-of-distribution data. Our experimental results outperform popular BNNs with fixed architectures and demonstrate comparable performance to deep ensembles but with significantly less runtime.
\end{itemize}

\section{Background}
\subsection{Bayesian deep learning}
It is well-known that an exact computation of the posterior distribution over model parameters of a modern Deep Neural Network is intractable. Bayesian Deep Learning methods rely on the mean-field assumption and Variational Bayes then allows us to find a distribution that approximates the true untractable posterior \cite{graves2011practical,kingma2013auto}.
\citeauthor{blundell2015weight} proposed Bayes-by-Backprop, and their stochastic weights can be reparameterized as deterministic and differentiable functions.
\citeauthor{gal2016dropout} used a a spike and slab variational distribution to reinterpret Dropout \cite{srivastava2014dropout} as approximate variational Bayesian inference.
\citeauthor{osawa2019practical} proposed to use natural-gradient Variational Inference to perform Bayesian Learning and demonstrated that Bayesian Learning is possible on larger datasets \cite{osawa2019practical}.
There are also recent advances in Bayesian Deep Learning that turns noisy optimization to Bayesian inference \cite{zhang2019cyclical,maddox2019simple}.

A typical Bayesian inference (using a BNN to run model inference) requires a Monte-Carlo estimate of the marginal likelihood of the posterior ($p(\hat{y}|\hat{x})$):

\begin{equation}
\mathE_{p(\hat{y}|\hat{x})} \approx \frac{1}{N} \sum_{n=0}^{N-1} f(\hat{x}, \btheta_n)
\end{equation}

$(\hat{x}, \hat{y})$ represents the test data point and $f$ is a neural network parameterized by $\btheta_n$. The above formulation requires $N$ samples and thus running model inference $N$ times. This formulation is the key component for providing us with the model uncertainty measurements but also introduces a significant runtime overhead. In addition, as these Bayesian Learning methods rely on the mean field assumption, this comes at the cost of reduced expressivity. In the field of model uncertainty measurements, canonical methods such as deep ensemble offer the best performance but also suffer from the above-mentioned inference overhead, in addition, ensemble models also have a significant training overhead \cite{lakshminarayanan2016simple}.

Another piece of work, Bayesian subnetwork inference \cite{daxberger2021bayesian}, shares a similar motivation to us; they suggest that a full BNN's uncertainty measurement capability might be well-preserved by a smaller sub-network from the original BNN. However, this proposed method does not provide any runtime reductions on GPUs as it still requires a Monte-Carlo based sampling over the full network as mentioned above.

Our NAS method is heavily inspired by Variational Inference based Bayesian Learning. In the experiments, we use Bayes-by-Backprop and MCDropout as the Bayesian baselines. More importantly, since the deep ensemble is seen as the state-of-the-art in model uncertainty calibration \cite{jospin2020hands}, with normally better or on par to Bayesian-based approaches, we will majorly focus on a comparison to it in our later evaluation.


\subsection{Network architecture search}

There are now two major directions in the field of NAS, namely Gradient-based and Evolutionary-based NAS methods.
Gradient-based NAS optimizes several pre-defined, trainable probabilistic priors \cite{liu2018darts,casale2019probabilistic,zhao2020probabilistic,zhao2020learned} and each scalar in these priors is associated with a pre-defined operation.
The probabilistic priors are then updated using standard Stochastic Gradient Descent.
Evolution-based NAS, on the other hand, operates on top of a pre-trained super-net and uses a search to rank the sub-network performances \cite{cai2018proxylessnas,cai2019once,zhao2021rapid}.
In this work, we follow the direction of Gradient-based NAS. We include a new formulation to optimize the network architectures using o.o.d data and also consider a new NAS optimization objective that is the uncertainty calibration.

Since the NAS problem can be viewed as a guided search that relies on prior observations, there is then a natural motivation to apply Bayesian Learning or Bayesian Optimization on NAS \cite{zhou2019bayesnas,white2019bananas}. Our implemented NAS algorithm, however, has a very different motivation from this line of work.
\section{Method}
\label{sec:method}

\begin{figure}[!t]
	\centering
	\includegraphics[width=.7\linewidth]{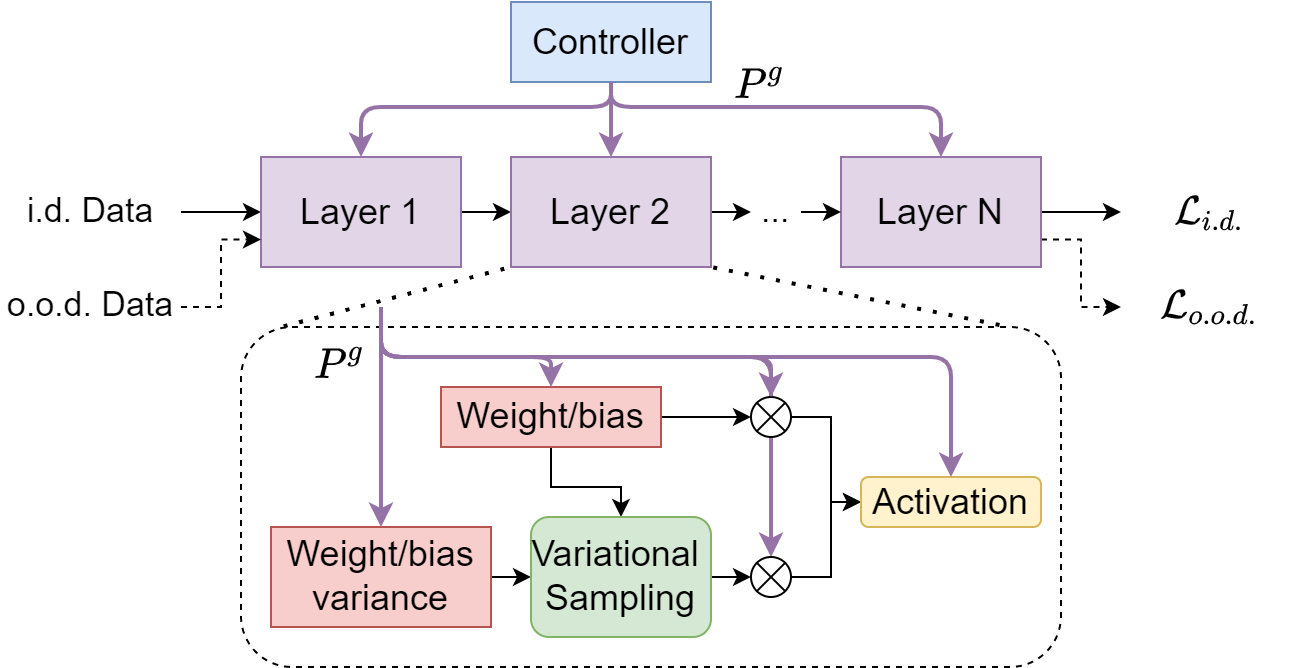}
	\caption{An overview of the Bayesian NAS framework. Controller outputs selection probability $P^g$ for candidate operators in each layer of the template network. For Bayesian layers, additional variance parameters are also selected with $P^g$ for variational sampling of weights. Controller selects the layer to be Bayesian or non-Baeysian with trainable scaling variables. The search is optimized both for the in-distribution (i.d.) data loss $\mathcal{L}_{i.d}$ and out-of-distribution (o.o.d) data loss $\mathcal{L}_{i.d}$ (Equation~\ref{eq:loss}).}
	\label{fig:overview}
\end{figure}

Figure~\ref{fig:overview} shows an overview of our Bayesian NAS framework. Our search framework contains a Bayesian template network (in purple) and a NAS controller (in blue). The Bayesian template network consists of stacks of layers of Bayesian and non-Bayesian candidate components. The choices of whether using a Bayesian or non-Bayesian layer, the activation functions, channel expansion counts, and kernel sizes together provide us a search space. 
During the search process, the NAS controller outputs selections of the candidate components of each layer in the template network. The layers with selected components are then assembled into a Bayesian model to process input data. 
We formulate the problem of searching for the best candidate network as a bi-level optimization, similar to typical gradient-based NAS strategies such as DARTS \cite{liu2018darts} using trainable scaling variables. However what differentiates our Bayesian NAS from other NAS methods is that \emph{we not only search in the classic NAS search space optimizing for in-distribution likelihood (usually categorical cross entropy for classification or Mean Squared Error for regression) but also have to search for the uncertainty performance using both in- and out-of-distribution data}. 
We propose a novel way of generating out-of-distribution data using Variational Auto-Encoders (VAEs)~\cite{kingma2013auto} for NAS algorithms. In the following subsections, we discuss the search space, the NAS controller, our out-of-distribution data generation, and the dual optimization objectives in detail.

\subsection{Search space}
Table~\ref{tab:search_space} illustrates a typical search space of each layer of the Bayesian template network for MNIST. 
We search over the expansion factors of the channel size (a.k.a number of hidden units), the type of activation functions, and whether the layer is a Bayesian layer. If the layer is a Bayesian layer, we use candidate weight parameters as the mean for variational sampling, and additionally, search over candidate matrices of weights' variances of the same shape as the weights' mean \cite{blundell2015weight}. For convolutional layers, we additionally search over the kernel size of the convolution kernels. Mathematically, we consider a Bayesian template network with $L$ layers and the number of possible search candidates in each layer is $S$.
For example, we have $S=4$ in \Cref{tab:search_space} ($4$ different search candidates). In a single layer presented in \Cref{tab:search_space}, the possible search options are $6 \times 6 \times 2 \times 3 = 216$ across these $4$ search candidates for convolutional layers and $72$ for linear layers (excluding the need of search for the kernel size).

For image classification tasks, we choose convolutional neural networks as our template networks. For MNIST dataset, we use a template network based on LeNet5~\cite{lecun1998gradient}. For CIFAR-10, our template network is ResNet18~\cite{he2016deep}.
The search spaces for MNIST and CIFAR10 then have slight differences due to GPU memory limits. 
The search space and details of these search backbones are described in \Cref{sec:appendix:backbone}.

The possible search space of the template network is large for even the simple LeNet5-based architectures containing 2 convolutional layers and 3 linear layers. This template network provides $216^{2}*72^{3} \approx 10^{8}$ possible combinations, which is infeasible to traverse manually.

\begin{table}[t!]
    \caption{
        Search space of each layer of the Bayesian template network for MNIST. Other search spaces are presented in our Appendix. ResNet18-based template has different channel expansion search options due to the limitation on GPU memory. 
    }
    \label{tab:search_space}
    \begin{center}
    \adjustbox{scale=0.75}{%
    \begin{tabular}{c|c}
    \toprule
    Search candidates & Possible options \\
    \midrule
    Channel Expansion & $0.5, 1, 1.5, 2, 4, 6, 8$ \\
    Activation functions & ReLU, ELU, SELU, Sigmoid, ReLU6, LeakyReLU \\
    Layer type & Non-bayesian, Bayesian \\
    Kernel Size & $1, 3, 5$ \\
    \bottomrule
    \end{tabular}
    }
    \end{center}
\end{table}

\subsection{NAS controller}
\label{sec:method:nas}
At each search iteration, the NAS Controller outputs selection probabilities $\bar{P}^g_{l,s}$ (named as $P^g$ in \Cref{fig:overview} for simplicity) for possible search candidates $s$ for the $l$th layer of the template network. 
The NAS controller is conditioned on trainable free variables $z$ such that it can model the joint distribution of the search options in all layers as $\bar{P}^g_{l,s} = P(l,s|z) P(z)$. In implementation we first embed $z$ with an Multi-Layer Perception (MLP) into $h(z)$. For $S$ possible search candidates and a network with $L$ layers, we then have to output $L \times S$ probability vectors. The details and dimensions of these MLPs are included in \Cref{sec:appendix:controller_mlp}, we use $f_{l,s}$ to represent another MLP that produces the output probability vector:

\begin{equation}
    \bar{P}^g_{l,s} = softmax(f_{l,k}(h(z)))
\end{equation}

\begin{figure}[!t]
	\centering
	\includegraphics[width=0.5\linewidth]{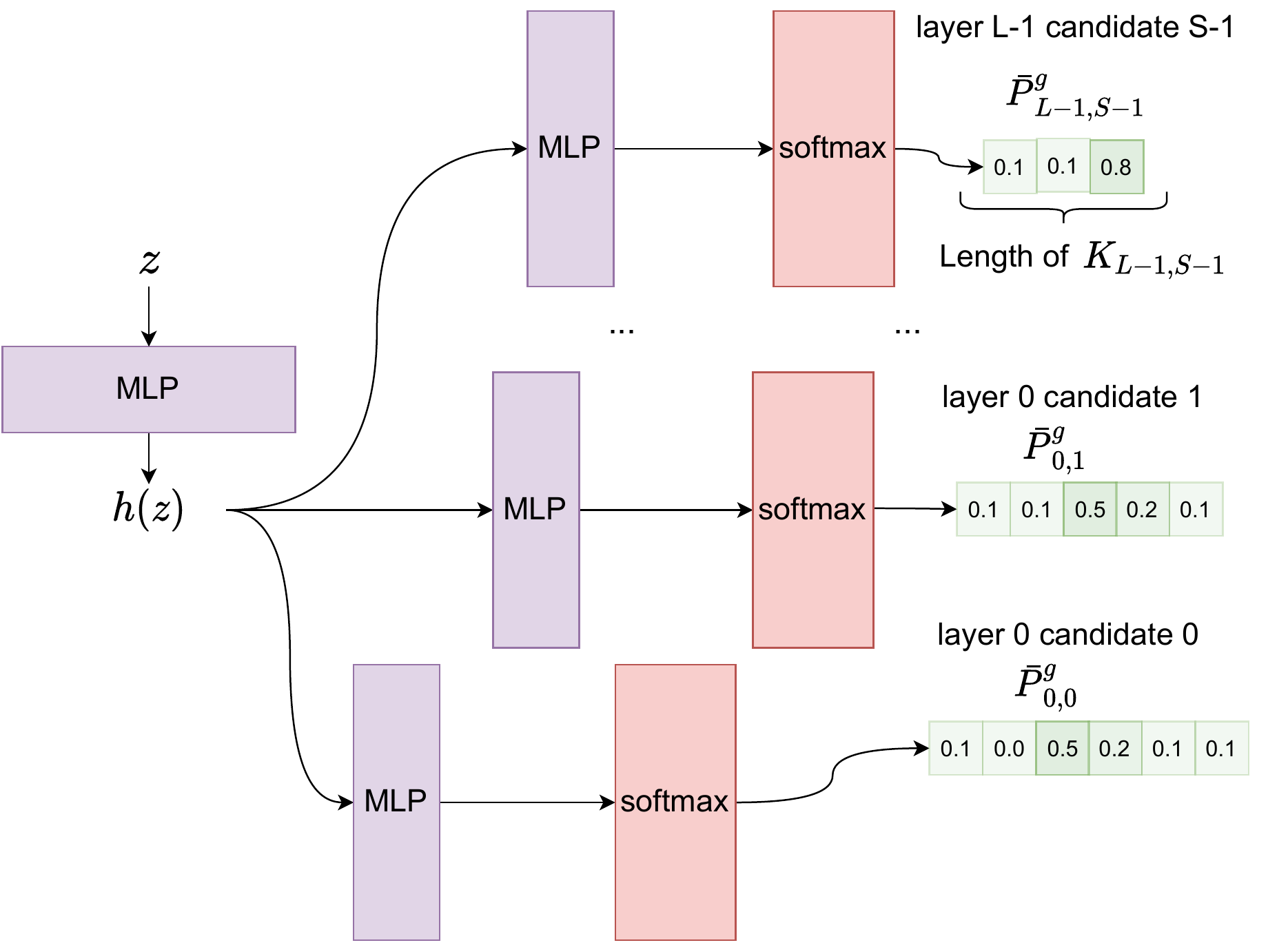}
	\caption{An overview of the NAS controller, where a trainable embedding $z$ produces a set of output probability vectors $\bar{P}^g_{l,s}$. $\bar{P}^g_{l,s}$ vectors are then utilized to pick candidate operations from the search space.}
	\label{fig:nas_controller}
\end{figure}
As illustrated in \Cref{fig:nas_controller}, the probability vector $\bar{P}^g_{l,s}$ has a  length of $K_{l,s}$ to represent selection probabilities for each of the $K_{l,s}$ candidates. To improve memory efficiency during the search, at each iteration we only select one candidate to be active for forward and backward pass, achieved via the $\mathsf{argmax}$ function.
This is also known as the single-path search strategy in previous NAS frameworks \cite{stamoulis2019single}.
By modeling the joint distribution $\bar{P}^g_{l,k}$ rather than the distribution for all possible $(l,s)$ pairs as $P(l,s)$, we further reduce the computational and memory complexity from $O(\sum_{l=0, s=0}^{L-1, S-1} K_{l, s}^{LS})$ to $O(\sum_{l=0, s=0}^{L-1, S-1}LSK_{l,s})$.

In experiments, we found that the probability distribution can occasionally greedily converge to poor local minima due to a lack of exploration of the search space. In order to force the controller to explore rather than exploiting the randomness of training, we use additional noise to encourage exploration. In practice, we found that noise with Gaussian distribution truncated to be both positive works well. We anneal the noise level linearly with a slope ($\lambda_n$) after a preset warmup epoch ($M_{w}$). We pick the parameters of the noise annealing based on a hyperparameter study on the MNIST dataset as shown in \Cref{sec:appendix:noise} ($\lambda_n = 0.1, M_w = 20$).

\subsection{Out-of-Distribution data generation}
\label{sec:method:ood}
In previous Bayesian learning work~\cite{hafner2020noise, daxberger2021bayesian}, out-of-distribution data is generated by directly adding noise or performing a randomized affine transformation on input data. These methods, while able to generate out-of-distribution data, have a limited degree of freedom. In this work, we propose to first learn a latent probability distribution of the data using VAEs, and add noise in the latent space to generate out-of-distribution data. In this way, the degree of freedom is as high as what the generator neural network can model. 

Typically, in a VAE, an encoder neural network embeds a datapoint $x$ to a latent distribution $N(\mu_x, \sigma_x)$ with a mean $\mu_x$ and a standard deviation $\sigma_x$. We then sample $z \sim N(\mu_x, \sigma_x)$ and the decoder network produces a reconstructed $x$ taken $z$ as an input \cite{kingma2013auto, kingma2019introduction}. In our formulation, we introduce another parameter $\beta$ so that the sampling occurs as $z \sim N(\mu_x, \beta + \sigma_x)$. In other words, $\beta$ controls the strength of the out-of-distribution data generation. A larger $\beta$ value means data generated from the VAE would be more `out-of-distribution'. 

\begin{figure}[!ht]
	\centering
	\includegraphics[width=.6\linewidth]{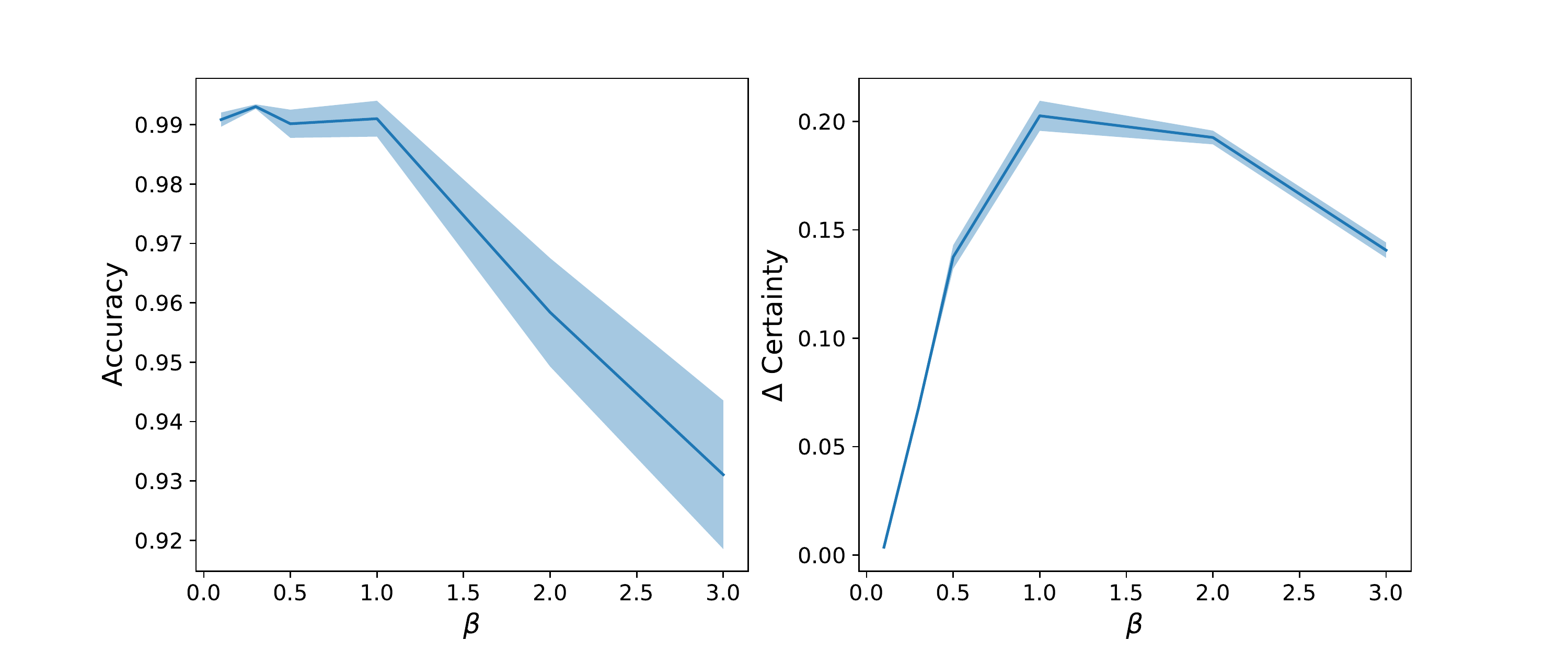}
	\caption{The effect of changing $\beta$ when generating the o.o.d data utilized in the proposed NAS algorithm. The first plot shows the accuracy of the searched model on MNIST and the second plot shows the differences of model prediction certainties between i.d (test data of MNIST) and o.o.d (FashionMNIST) data. Results are averaged across three independent runs.}
	\label{fig:noise_std}
\end{figure}
\begin{table}[ht]
	\caption{The effect of VAE generated o.o.d data on MNIST.
	$\beta=0.0$ would generate i.d data for NAS optimization.
	Both Without VAE and $\beta=0.0$ 
	give limited uncertainty quantifying abilities, \ie~ the model cannot distinguish between i.d (MNIST data) and o.o.d data (FashionMNIST) with a small $\Delta$ Certainty gap (bigger is better).
	}
	\label{tab:ood}
	\begin{center}
	\adjustbox{scale=0.8}{%
	\begin{tabular}{c|ccc}
	\toprule
	Method   & Without VAE	& With VAE 	&  With VAE \\
	Params 	& & $\beta = 0$ 	&  $\beta = 1.0$\\
	\midrule
	Accuracy 
	& 0.9923	& 0.9908 	& 0.9910 \\
	$\Delta$ Certainty	
	& 0.0421 	& 0.0036	& 0.2026 \\
	\bottomrule
	\end{tabular}
	}
	\end{center}
\end{table}

The network architectures and training strategies of our VAEs used on different datasets are included in our \Cref{sec:appendix:vae}. 
\Cref{fig:noise_std} indicates a trade-off of different $\beta$ values. A larger $\beta$ value is often beneficial for the model to gain prediction certainty gaps between in- and out-of-distribution data. This means the model shows better performance and knows to be uncertain about data that is out-of-distribution. In the meantime, a large $\beta$ value can hurt classification accuracy on in-distribution data. Based on the results in \Cref{fig:noise_std}, we pick $\beta = 1.0$ for a balance between accuracy and certainty difference and use it for the rest of our evaluation.

To further demonstrate the effectiveness of using correctly generated o.o.d data in NAS, we present in \Cref{tab:ood} how the NAS performs without the VAE generated data, with incorrectly generated VAE data ($\beta = 0$) and a well-tuned o.o.d data ($\beta=1.0$). Although all three approaches show similar accuracy metrics because of the NAS algorithm, well-tuned o.o.d data generation brings a significant performance boost when looking at the uncertainty measurements. 
Intuitively, when $\beta = 0$, the VAE generates i.d data for the NAS and does not contribute to uncertainty calibration.

\subsection{Optimization}\label{sec:optimisation}
\label{sec:method:opt}
We formulate NAS as a bi-level optimization problem
similar to DARTS \cite{liu2018darts}:
\begin{equation}
\begin{split}
\min_{a} & \, \mathcal{L}_{val} (\theta^\star(a),a) \\
\text{s.t. } & \, \theta^\star(a) = \argmin_{\theta} (\mathcal{L}_{train}(\theta,a))
\end{split}
\end{equation}
Here, $\theta$ are the parameters of all candidate operators,
$\theta^\star(a)$ is the optimal parameters given $a$, where $a$ represents
parameters of the architecture search controller $a$.
$L_{train}$  is a training loss on the training data split,
while $L_{val}$ is validation loss on the validation data split.
The parameters $\theta$ and $a$ are trained iteratively
with their own optimizers.
Since it is computationally intractable to compute $\theta^\star(a)$
for each update of $a$,
we approximate $\theta^\star$ with a few training steps,
which are shown to be effective in DARTS~\cite{liu2018darts}.

For the training loss objective $\mathcal{L}_{train}$, we use corresponding likelihood objectives for different tasks (cross entropy for classification while MSE loss for regression). For the validation loss, in addition to the likelihood objective, we also include variance objectives for both the in-distribution and out-of-distribution data. The approximation of these variance objectives (using a variational distribution) is the same as the one utilized in \citeauthor{blundell2015weight}, and a detailed discussion can be found in \citeauthor{jospin2020hands}. We minimize the validation set variances for in-distribution (i.d) data and maximize the variance for out-of-distribution (o.o.d) data (negative sign, and as illustrated in \Cref{fig:overview}). Therefore the validation loss becomes:
\begin{equation}\label{eq:loss}
\mathcal{L}_{val} = \underbrace{\mathcal{L}_{likelihood} + \alpha \widehat{Var}_{i.d}}_{\mathcal{L}_{i.d}} -  \underbrace{\gamma \widehat{Var}_{o.o.d}}_{\mathcal{L}_{o.o.d}}
\end{equation}
Where $\mathcal{L}_{i.d}$ comprises likelihood and variance for i.d data, while $\mathcal{L}_{o.o.d}$ is the variance for o.o.d data. 
We consider $\alpha \in \{0.0001, 0.001, 0.01, 0.1\}$ and $\gamma \in \{0.0001, 0.001, 0.01, 0.1\}$. In practice we pick the $\alpha$ and $\gamma$ values using only one epoch of training. We would like to make sure $\alpha \widehat{Var}_{i.d}$, $\gamma \widehat{Var}_{o.o.d}$ and $\mathcal{L}_{likelihood}$ are on the same level of magnitudes after an epoch of searching, so there is no overly dominant loss term that off-balances the optimization.
In the implementation we found setting both $\alpha$ and $\gamma$ to $0.01$ often satisfies the above requirement, this type of control of $\alpha$ is also discussed in prior work on training Bayesian Neural Networks \cite{blundell2015weight} (a.k.a KL re-weighting).
\section{Experiments}
\label{sec:exp:overview}
\begin{table*}[ht!]
	\caption{
		MNIST, MNIST rotated, MNIST corrupted and FashionMNIST.
		$\Delta$ Certainty (bigger is better) shows the certainty gap compared to in-distribution data.
		NLL is the negative log-likelihood.
		Results are averaged across three independent runs.
	}
	\label{tab:mnist}
	\begin{center}
	\adjustbox{scale=.85}{%
	\begin{tabular}{c|ccccc}
	\toprule
	\multirow{1}*{Method}   
	& Non-Bayesian	& LRT	& MCDropout	& Ensemble	& NAS \\
	\midrule
	& \multicolumn{5}{c}{MNIST (In distribution data)} \\
	\midrule
	Accuracy 
	& $\mathit{0.9902 \pm 0.0008}$
	& $0.9896 \pm 0.0005$ 
	& $0.9342 \pm 0.0061$ 
	& $0.9864 \pm 0.0002$ 
	& $\mathbf{0.9910 \pm 0.0011}$
	\\
	Certainty
	& $0.9984 \pm 0.0003$
	& $0.9988 \pm 0.0001$
	& $0.8965 \pm 0.0585$ 
	& $0.9803 \pm 0.0008$ 
	& $0.9951 \pm 0.0020$
	\\
	NLL
	& $0.2103 \pm 0.0260$ 
	& $0.1001 \pm 0.0014$ 
	& $0.7887 \pm 0.0632$ 
	& $0.0458 \pm 0.0018$ 
	&$0.0375 \pm 0.0062$
	\\
	\midrule
	& \multicolumn{5}{c}{MNIST rotated 30 degrees (Out of distribution data)}\\
	\midrule
	Accuracy 
	&$0.8663 \pm 0.0102$ 
	&$0.8753 \pm 0.0107$ 
	&$0.5954 \pm 0.0352$ 
	&$0.8482 \pm 0.0031$
	&$0.8545 \pm 0.0059$
	\\
	Certainty 
	&$0.9752 \pm 0.0032$ 
	&$0.9874 \pm 0.0008$ 
	&$0.8896 \pm 0.0617$ 
	&$0.8516 \pm 0.0030$ 
	&$0.9475 \pm 0.0164$
	\\
	NLL
	&$3.4135 \pm 0.1755$ 
	&$1.6534 \pm 0.1239$ 
	&$22.9911 \pm 12.5153$ 
	&$0.5163 \pm 0.0085$ 
	&$0.7885 \pm 0.1380$
	\\
	$\Delta$ Certainty 
	&$0.0232$	
	&$0.0114$	
	&$0.0068$	
	&$\mathbf{0.1287}$  
	&$\mathit{0.0476}$\\
	\midrule
	& \multicolumn{5}{c}{MNIST corrupted level 1 (Out of distribution data)}\\
	\midrule
	Accuracy 
	&$0.0992 \pm 0.0039$ 
	&$0.0957 \pm 0.0028$ 
	&$0.1066 \pm 0.0038$ 
	&$0.1129 \pm 0.0009$ 
	&$0.1009 \pm 0.0016$
	\\
	Certainty 
	&$0.9755 \pm 0.0254$ 
	&$0.8507 \pm 0.0435$ 
	&$0.8334 \pm 0.0736$ 
	&$0.7996 \pm 0.0496$ 
	&$0.8438 \pm 0.0827$
	\\
	NLL
	&$10.0627 \pm 0.8233$ 
	&$26.0908 \pm 14.2407$ 
	&$34.0331 \pm 14.2999$ 
	&$7.6489 \pm 0.0380$ 
	&$13.6517 \pm 5.3868$
	\\
	$\Delta$ Certainty 
	&$0.0147$	
	&$0.1481$	
	&$0.0630$	
	&$\mathbf{0.1807}$
	&$\mathit{0.1513}$\\
	\midrule
	& \multicolumn{5}{c}{FashionMNIST (Out of distribution data)}\\
	\midrule
	Accuracy 
	&$0.0986 \pm 0.0018$ 
	&$0.0973 \pm 0.0006$ 
	&$0.1040 \pm 0.0007$ 
	&$0.1058 \pm 0.0007$ 
	&$0.0923 \pm 0.0256$
	\\
	Certainty 
	&$0.9668 \pm 0.0331$ 
	&$0.9538 \pm 0.0422$ 
	&$0.8546 \pm 0.0753$ 
	&$0.7011 \pm 0.0396$ 
	&$0.7925 \pm 0.0503$
	\\
	NLL 
	&$15.0814 \pm 1.4024$ 
	&$32.1182 \pm 16.4905$ 
	&$54.6312 \pm 24.9626$ 
	&$5.4911 \pm 0.2160$ 
	&$12.6002 \pm 3.6958$
	\\
	$\Delta$ Certainty 
	&$0.0316$	
	&$0.0450$	
	&$0.0418$	
	&$\mathbf{0.2792}$  
	&$\mathit{0.2026}$\\
	\midrule
	& \multicolumn{5}{c}{Inference Time with a batch size of 128 on NVIDIA GeForce RTX 2080 Ti (ms)}\\
	\midrule
	Latency
	& $\mathbf{12.0341 \pm 0.5534}$
	& $155.78073 \pm 5.2058$ 
	& $102.7167 \pm 2.0197$
	& $107.5535 \pm 6.7008$
	& $\mathit{53.8164 \pm 0.2321}$
	\\
	\bottomrule
	\end{tabular}
	}
	\end{center}
	\vskip -10pt
    \end{table*}

We present our results with the following baselines on two image classification tasks, a loan approval prediction task and a heart disease prediction task.
\begin{itemize}
    \item Point estimates or Non-Bayesian Neural Networks (Non-Bayesian): models trained with canonical Stochastic Gradient Descent and the network's post-softmax outputs are used as a certainty metric.
    \item Bayes-by-Backprop networks with the local reparameterization trick (LRT): these models use unbiased Monte Carlo estimates to update the gradients \cite{blundell2015weight}.  
    \item Monte-Carlo Dropout (MCDropout): this method makes the use of the Dropout layer \cite{srivastava2014dropout} to provide uncertainty measurements, where it can be interpreted as a less expressive inference compared to the Bayes-by-Backprop networks \cite{gal2016dropout}.
    \item Deep Model Ensemble (Ensemble): this method does not use Bayesian inference but is normally seen as the state-of-the-art in uncertainty measurements \cite{lakshminarayanan2016simple}.    
\end{itemize}

For all the baselines, we pick the best performing models trained with a set of learning rates $\{1e^{-2}, 1e^{-3}, 1e^{-4}\}$ and use a standard Adam optimizer \cite{kingma2014adam}. For any methods requiring a Monte-Carlo sampling, we use $10$ samples for each approximation. In our experiment, we make sure each Deep Ensemble contains $10$ models. In this case, the inference runtime of all baselines would be similar since all methods would have to perform $10$ inference runs. 

\subsection{Image classification with distribution shifts}
\label{sec:eval:image}
We consider two image classification tasks: MNIST \cite{deng2012mnist} and CIFAR10 \cite{krizhevsky2009learning}, and include o.o.d datasets to examine the performance of the models with data distribution shifts:

\begin{itemize}
    \item MNIST rotated 30 degrees: it contains all images in the test set of MNIST rotated by 30 degrees. Rotate 30 degrees is only considered as an o.o.d dataset for MNIST but not CIFAR10, because the training of the CIFAR10 model utilizes a random rotation augmentation. 
    \item MNIST/CIFAR10 corrupted level $n$: we use corruptions introduced by \citeauthor{michaelis2019benchmarking} with their openly available library. The corruptions are divided into $n$ levels, where the higher $n$ values indicate a more severe corruption. The corruptions are applied on the test partition of the datasets.
    \item FashionMNIST: Fashion-MNIST is a dataset of Zalando's images, we use its test set that contains 10,000 examples to serve as an o.o.d dataset for MNIST \cite{xiao2017fashion}.
    \item Street View House Numbers (SVHN): SVHN is a real-world image dataset for recognizing digits and numbers in natural scene images, we use its  test set as an o.o.d dataset for CIFAR10 \cite{netzer2011reading}.
\end{itemize}

We use the hyper-parameters decided from \Cref{sec:method}, which we also summarize again here: $M_{w} = 20$, $\lambda_n = 0.1$ (\Cref{sec:method:nas}); $\beta=1.0$ (\Cref{sec:method:ood}); $\alpha = \sigma = 0.01$ (\Cref{sec:method:opt}). There are also two learning rates, one for training the Bayesian template network and one for the NAS controller. Similar as the treatment to the baselines, we manually select the learning rates from  $\{1e^{-2}, 1e^{-3}, 1e^{-4}\}$ (a discussion about learning rate selections is in \Cref{sec:appendix:lr}).   

\begin{table*}[ht!]
	\caption{
		CIFAR10, CIFAR10 corrupted and SVHN.
		$\Delta$ Certainty (bigger is better) shows the certainty gap compared to in-distribution data.
		NLL is the negative log-likelihood.
		Results are averaged across three independent runs.
	}
	\label{tab:cifar10}
	\begin{center}
	\adjustbox{scale=0.85}{%
	\begin{tabular}{c|ccccc}
	\toprule
	\multirow{1}*{Method}   
	& Non-Bayesian	& LRT	& MCDropout	& Ensemble	& NAS \\
	\midrule
	& \multicolumn{5}{c}{CIFAR10 (In distribution data)} \\
	\midrule
	Accuracy 
	&   $\mathit{0.9184 \pm 0.0008}$       
	&   $\mathbf{0.9187 \pm 0.0043}$ 
	&   $0.8708 \pm 0.0110$ 
	&   $0.9176 \pm 0.0011$ 
	&   $0.9143 \pm 0.0233$ 
	\\
	Certainty
	&   $0.9922 \pm 0.0004$ 
	&   $0.9921 \pm 0.0004$ 
	&   $0.9671 \pm 0.0013$ 
	&   $0.9921 \pm 0.0001$ 
	&   $0.9843 \pm 0.0011$ 
	\\
	NLL
	&   $0.7544 \pm 0.0240$ 
	&   $0.6541 \pm 0.0449$ 
	&   $1.2205 \pm 0.1253$ 
	&   $0.7304 \pm 0.0161$ 
	&   $0.7422 \pm 0.0203$ 
	\\
	\midrule
	& \multicolumn{5}{c}{CIFAR10 corrupted leve 1 (Out of distribution data)}\\
	\midrule
	Accuracy 
	&$0.1098 \pm 0.0086$ 
	&$0.1116 \pm 0.0072$ 
	&$0.0977 \pm 0.0019$ 
	&$0.1124 \pm 0.0052$ 
	&$0.1011 \pm 0.0079$
	\\
	Certainty 
	&$0.8829 \pm 0.0707$ 
	&$0.8772 \pm 0.0889$ 
	&$0.9076 \pm 0.0125$ 
	&$0.8487 \pm 0.0077$ 
	&$0.7899 \pm 0.0171$
	\\
	NLL
	&$15.8335 \pm 7.2911$ 
	&$14.7672 \pm 2.8791$ 
	&$18.8109 \pm 2.5662$ 
	&$14.0338 \pm 1.1402$ 
	&$14.7175 \pm 1.0371$
	\\
	$\Delta$ Certainty 
	&$0.1092$	
	&$0.1150$	
	&$0.0595$	
	&$\mathit{0.1434}$ 
	&$\mathbf{0.1944}$ 
	\\
	\midrule
	& \multicolumn{5}{c}{CIFAR10 corrupted level 3 (Out of distribution data)}\\
	\midrule
	Accuracy 
	&$0.0999 \pm 0.0026$ 
	&$0.1005 \pm 0.0026$ 
	&$0.0987 \pm 0.0017$ 
	&$0.1009 \pm 0.0006$ 
	&$0.1121 \pm 0.0020$
	\\
	Certainty 
	&$0.8898 \pm 0.1013$ 
	&$0.8493 \pm 0.0773$ 
	&$0.9090 \pm 0.0382$ 
	&$0.8349 \pm 0.0463$ 
	&$0.7864 \pm 0.0249$
	\\
	NLL
	&$15.9620 \pm 13.1737$ 
	&$13.6133 \pm 3.3533$ 
	&$18.2715 \pm 3.9940$ 
	&$11.9185 \pm 0.9514$ 
	&$12.4533 \pm 0.8312$ 
	\\
	$\Delta$ Certainty 
	&$0.1024$	
	&$0.1428$	
	&$0.0581$	
	&$\mathit{0.1573}$ 
	&$\mathbf{0.1979}$ 
	\\
	\midrule
	& \multicolumn{5}{c}{SVHN (Out of distribution data)}\\
	\midrule
	Accuracy 
	&$0.1040 \pm 0.0038$ 
	&$0.1128 \pm 0.0017$ 
	&$0.1263 \pm 0.0033$ 
	&$0.1094 \pm 0.0017$
	&$0.1025 \pm 0.0087$
	\\
	Certainty 
	&$0.9469 \pm 0.0111$ 
	&$0.9437 \pm 0.0124$ 
	&$0.9376 \pm 0.0111$ 
	&$0.9477 \pm 0.0026$
	&$0.7960 \pm 0.0134$
	\\
	NLL 
	&$24.4307 \pm 2.7540$ 
	&$23.8798 \pm 3.6388$ 
	&$22.2501 \pm 2.2782$ 
	&$26.6406 \pm 1.3254$ 
	&$20.0216 \pm 1.0278$ 
	\\
	$\Delta$ Certainty 
	& $0.0453$
	& $\mathit{0.0484}$ 
	& $0.0295$ 
	& $0.0444$ 
	& $\mathbf{0.1883}$ 
	\\
	\midrule
	& \multicolumn{5}{c}{Inference Time with a batch size of 32 on NVIDIA GeForce RTX 2080 Ti (ms)}\\
	\midrule
	Latency & $\mathbf{10.9965 \pm 0.4044}$
	& $188.1846 \pm 3.34016$
	& $102.1346 \pm 6.6028$
	& $104.0347 \pm 0.5535$ 
	& $\mathit{34.9785 \pm 1.3234}$ 
	\\
	\bottomrule
	\end{tabular}
	}
	\end{center}
	\vspace{-20pt}
    \end{table*}

\Cref{tab:mnist} shows how the proposed NAS algorithm performs compared to the baselines. the $\Delta$ Certainty row shows the difference in the certainty metrics between i.d and o.o.d data. 
All results are from three independent search and retrain runs.
It is worth pointing out that during the search of the architecture, our algorithm has also never seen any of these o.o.d datasets listed in \Cref{tab:mnist}.
The NAS algorithm shows the best accuracy on i.d data, in addition, it achieves the second-best uncertainty performance in \Cref{tab:mnist} (\ie~has the second-best score in $\Delta$ Certainty).

On the CIFAR10 dataset, in \Cref{tab:cifar10}, the NAS algorithm shows comparable accuracy to a range of baseline methods. Surprisingly, it then also demonstrates the best ability in quantifying uncertainty. We hypothesize the deep ensemble approach shows less dominant numbers on this dataset because of the increased task complexity -- the number of models in the ensemble might have to increase with the increased task complexity. However, in our comparison, to make sure the deep ensemble will have a similar runtime to BNNs, we kept the number of models in a deep ensemble to be $10$. In contrast, our search then has the ability to automatically adapt to increased difficulties by exploring the architectural design space and therefore shows the best uncertainty measurements.

\begin{table*}[ht!]
    \caption{
        Results of Heart Disease UCI.
        The objective is to predict the presence of heart disease in the patient.
        $\Delta$ Certainty (bigger is better) shows the certainty metric gap between in- and out-of-distribution data.
		NLL is the negative log-likelihood.
		Results are averaged across three independent runs.
    }
    \label{tab:heart}
    \begin{center}
    \adjustbox{scale=0.9}{%
    \begin{tabular}{c|ccccc}
    \toprule
    \multirow{1}*{Method}   
    & F1 Score    &AUROC    & Certainty & $\Delta$ Certainty    & NLL  \\
    \midrule
    & \multicolumn{5}{c}{In distribution data} \\
    \midrule
    Non-Bayesian	
    &$0.9785 \pm 0.0152$	
    &$0.9792 \pm 0.0147$	
    &$0.9893 \pm 0.0100$	
    & - &$0.0821 \pm 0.0771$\\
    LRT	
    &$0.9375 \pm 0.0000$	
    &$0.9375 \pm 0.0000$	
    &$0.9514 \pm 0.0028$	
    & - &$0.1316 \pm 0.0132$\\
    MCDropout	
    &$0.9882 \pm 0.0112$	
    &$0.9877 \pm 0.0117$	
    &$0.9919 \pm 0.0040$	
    &-  &$0.0377 \pm 0.0454$\\
    Ensemble	
    &$0.9892 \pm 0.0152$	
    &$0.9896 \pm 0.0147$	
    &$0.9913 \pm 0.0029$	
    &-  &$0.0649 \pm 0.0502$\\
    NAS 
    &$\mathbf{0.9936 \pm 0.0013}$  
    &$\mathbf{0.9918 \pm 0.0230}$ 
    &$0.9977 \pm 0.0033$   & - &$0.0000 \pm 0.0000$  \\
    \midrule
    & \multicolumn{5}{c}{Out of distribution data}\\
    \midrule
    Non-Bayesian	
    &$0.6628 \pm 0.0533$	
    &$0.5143 \pm 0.0157$	
    &$0.9074 \pm 0.0539$	
    & $0.0819$
    &$1.6745 \pm 0.3862$\\
    LRT	
    &$0.7697 \pm 0.0087$	
    &$0.5212 \pm 0.0027$	
    &$0.6982 \pm 0.0077$	
    & $0.1532$
    &$0.7795 \pm 0.0112$\\
    MCDropout	
    &$0.6475 \pm 0.0378$	
    &$0.5198 \pm 0.0184$	
    &$0.8568 \pm 0.0197$	
    & $0.1351$
    &$1.3935 \pm 0.1628$\\
    Ensemble	
    &$0.6735 \pm 0.0170$	
    &$0.5171 \pm 0.0059$	
    &$0.9186 \pm 0.0037$	
    & $0.1727$
    &$1.8723 \pm 0.0336$\\
    NAS 
    &$0.5709 \pm 0.1056$  
    &$0.5109 \pm 0.0118$  
    &$0.8231 \pm 0.1952$  
    & $\mathbf{0.1746}$
    &$9.0560 \pm 2.1621$  \\
    \bottomrule
    \end{tabular}
    }
    \end{center}
    \vspace{-15pt}
\end{table*}

\begin{table*}[ht!]
    \caption{
        Results of Loan Approval.
        The objective is to predict the whether a house loan is approved or not.
        $\Delta$ Certainty (the bigger the better) shows the certainty metric gap between in- and out-of-distribution data.
        NLL is the negative log-likelihood.
		Results are averaged across three independent runs.
    }
    \label{tab:loan}
    \begin{center}
    \adjustbox{scale=0.9}{%
    \begin{tabular}{c|ccccc}
    \toprule
    \multirow{1}*{Method}   
    & F1 Score    &AUROC & Certainty & $\Delta$ Certainty & NLL  \\
    \midrule
    & \multicolumn{5}{c}{In distribution data} \\
    \midrule
    Non-Bayesian	
    &$0.8087 \pm 0.0070$	
    &$0.6365 \pm 0.0121$	
    &$0.9222 \pm 0.0073$	
    & - &$1.3513 \pm 0.0500$\\
    LRT	
    &$0.8146 \pm 0.0097$	
    &$0.5294 \pm 0.0415$	
    &$0.8031 \pm 0.0173$	
    & -
    &$0.6116 \pm 0.0025$\\
    MCDropout	
    &$0.7679 \pm 0.0204$	
    &$0.5540 \pm 0.0263$	
    &$0.9079 \pm 0.0036$	
    & -
    &$1.4507 \pm 0.1558$\\
    Ensemble	
    &$\mathbf{0.8454 \pm 0.0000}$	
    &$0.6381 \pm 0.0000$	
    &$0.8024 \pm 0.0040$	
    & -
    &$0.5694 \pm 0.0029$\\
    NAS 
    &$0.8298 \pm 0.0074$  
    &$\mathbf{0.6393 \pm 0.0006}$ 
    &$0.8715 \pm 0.0022$   & - &$0.6914 \pm 0.1203$ \\
    \midrule
    & \multicolumn{5}{c}{Out of distribution data}\\
    \midrule
    Non-Bayesian	
    &$0.6041 \pm 0.0794$	
    &$0.5067 \pm 0.0111$	
    &$0.8891 \pm 0.0897$	
    &$0.0331$
    &$2.1252 \pm 0.8270$\\
    LRT	
    &$0.6667 \pm 0.0000$	
    &$0.5000 \pm 0.0000$	
    &$0.7200 \pm 0.0261$	
    &$0.0831$
    &$0.8143 \pm 0.0373$\\
    MCDropout	
    &$0.5706 \pm 0.0990$	
    &$0.4907 \pm 0.0090$	
    &$0.8950 \pm 0.0724$	
    &$0.0831$
    &$2.1458 \pm 1.0629$\\
    Ensemble	
    &$0.6667 \pm 0.0000$	
    &$0.5000 \pm 0.0000$	
    &$0.7425 \pm 0.0137$	
    &$0.1599$
    &$0.8553 \pm 0.0187$\\
    NAS 
    &$0.6332 \pm 0.0074$  
    &$0.5032 \pm 0.0021$ 
    &$0.6879 \pm 0.0310$   
    & $\mathbf{0.1836}$ 
    &$0.7679 \pm 0.0201$ \\
    \bottomrule
    \end{tabular}
    }
    \end{center}
    \vspace{-15pt}
\end{table*}
\subsection{Inference efficiency}
\label{sec:eval:efficiency}
\Cref{tab:mnist} and \Cref{tab:cifar10} also have a row that show the inference time of running these models.
It is worth mentioning that the inference time is averaged across $300$ runs to provide a faithful reading.

The proposed NAS algorithm assigns layers to be Bayesian from the last layer to the first. This means the network will always try to assign later layers in the network to be Bayesian \cite{brosse2020last,zeng2018relevance}.
The intuition is that it might be redundant to perform uncertainty measurements using the entire BNN, which is also empirically observed in \citeauthor{daxberger2021bayesian}. Bayesian inference on the last few layers might be sufficient for uncertainty calibration.
For instance, if a $5$-layer network is selected to have $2$ Bayesian layers, this means only the last $2$ layers are Bayesian. In this way, we have the ability to freeze the intermediate results after computing the first $3$ layers, the later $2$ layers can then use Bayesian inference to build a posterior estimation for the output.
As a result, both of \Cref{tab:mnist} and \Cref{tab:cifar10} show that the searched models are significantly faster than the baselines that have all layers being Bayesian. 

\begin{figure}[!h]
	\centering
	\includegraphics[width=.6\linewidth]{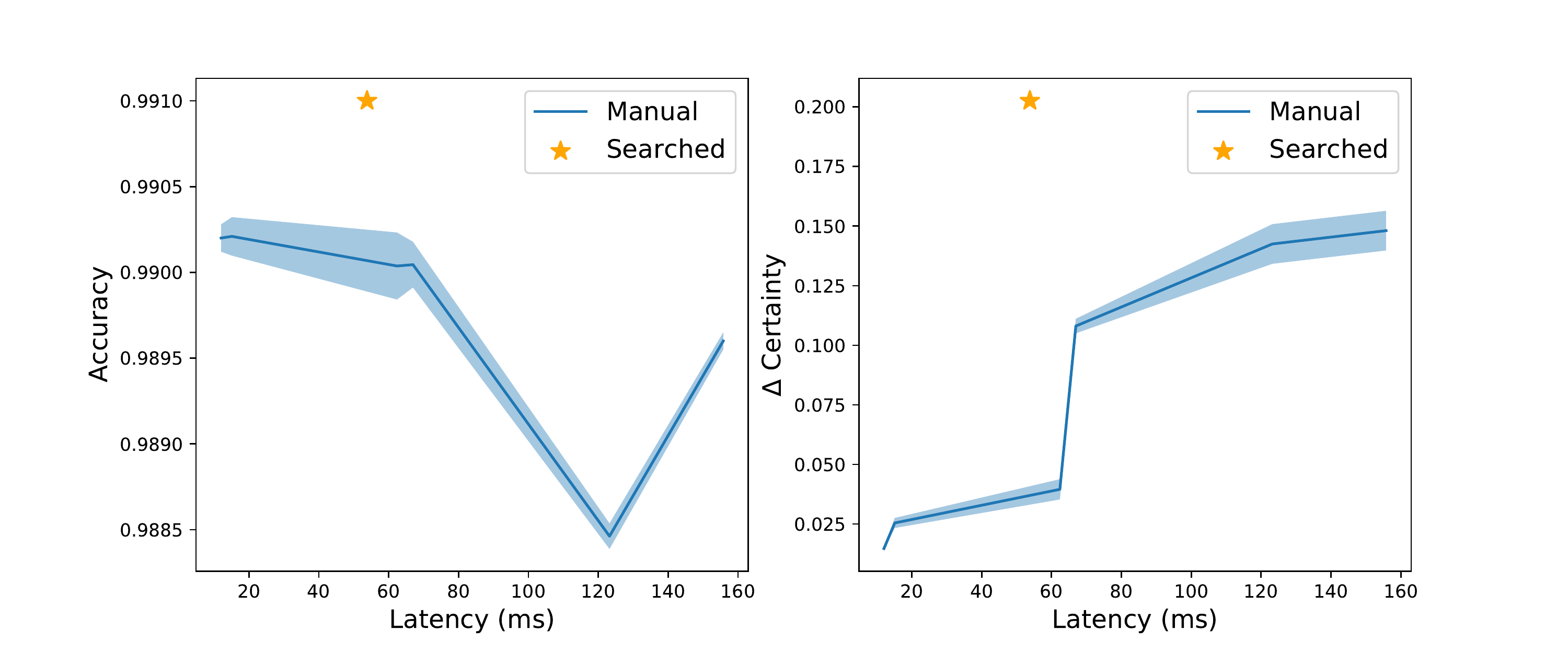}
	\caption{Latency \vs Accuracy and Latency \vs $\Delta$ Certainty tradeoffs. The blue line shows a Bayesian Neural Network, trained with LRT, with Bayesian inference on the ($n$-)last layer(s) only for MNIST classification. The searched network (orange star) greatly outperforms the manual baseline. Results are averaged across three independent runs and latency is measured on NVIDIA GeForce RTX 2080 Ti GPUs with a batch size of 128.}
	\label{fig:speed}
\end{figure}

Prior work has also demonstrated the use of Bayesian inference only on the ($n$-)last layer(s) only, intuitively, this approach can be seen as learning a point estimate transformation followed by a shallow BNN \cite{jospin2020hands}. The user, however, has to determine which $n$ last layer(s) is/are Bayesian by hand on a fixed architecture.
\Cref{fig:speed} shows how this manually designed last layer(s) only Bayesian approach (blue line) compares to our searched method (orange dot). The elaborated search space helps NAS to outperform the ($n$-)last layer(s) baselines by a significant margin. We present more data and detail  in \Cref{sec:appendix:inference}.

\subsection{Potential applications with uncertainty measurements}
In this section, we consider more realistic classification tasks (\eg~medical and financial tasks) that might be beneficial from having an uncertainty measurement. In these tasks, prevention of rare yet costly mistakes should normally be provided,
in particular, we consider:
\begin{itemize}
    \item Heart Disease HCI: This database contains 14 attributes to describe the conditions of a patient, the task is to classify whether the patient has potential heart diseases, this dataset is from the UC Irvine Machine Learning Repository \cite{asuncion2007uci}. All features are manually normalized.
    \item Loan Approval Prediction: This dataset contains 12 useful features about a customer applying for a house loan, the task is to predict whether the applicant is eligible for the loan \cite{loan}. 
    All features are manually normalized.
    \item Out of distribution data: we generate o.o.d data for these datasets by using random features (white noise).
\end{itemize}

In both \Cref{tab:heart} and \Cref{tab:loan}, our NAS generates results with high F1 scores (the best in Heart Disease HCI and the second-best in Loan Approval). In addition, the searched models show the best $\Delta$ Certainty on both datasets. The details about the network architectures for both the baselines and our search template model are in \Cref{sec:appendix:backbone}.
Inference time is not discussed in this case, since the runtime on small datasets is not likely to be a bottleneck in today's systems.

On these simpler classification tasks, we see that our NAS algorithm outperforms the deep ensemble. We hypothesize that NAS algorithms are more advantageous on simple datasets. 
Intuitively, since these two tasks are significantly easier, the NAS method now contains a search space that has more possible models that will perform well. The NAS becomes an easier problem because there are now more equivalently good architectures for the algorithm to converge to. 

On the other hand, we use \Cref{tab:heart} and \Cref{tab:loan} to demonstrate the effectiveness of the NAS method on applications that involve critical decision making and how NAS algorithm can help BNNs to achieve significantly better performance compared to BNNs with fixed architectures.

\section{Conclusion}

In this paper, we demonstrate a network architecture search method that can help Bayesian Neural Networks to find a suitable network architecture based on the targeting dataset. The proposed NAS method searches for architectures with not only the best accuracy but also a well-tuned certainty metric. The proposed method, unlike existing NAS approaches, makes use of both i.d and o.o.d data to achieve its optimization targets.

Empirically, we demonstrate that our NAS method can achieve comparable accuracy and uncertainty calibration compared to the deep ensemble on MNIST and CIFAR10. More importantly, the searched model can reduce the runtime by around $3 \times$ compared to various BNN baselines and the deep ensemble. 
\newpage
\appendix
\section{Baseline networks, NAS backbones and their search spaces}
\label{sec:appendix:backbone}
In \Cref{tab:search_space}, we illustrate a typical search space for MNIST. The only component of the search space that is different are the expansion factors. This is mainly because of the GPU memory limitation. For CIFAR10, the expansion factors are $\{0.5, 1, 1.5, 2, 2.5, 3.0, 3.5\}$ to ensure there is no Out-of-Memory error.

The Bayesian template network can have different structures. We use a LeNet5 based structure for problems on MNIST \cite{lecun1998gradient}, and the ResNet-based structures \cite{he2016deep} is used on CIFAR10 classification.
\Cref{tab:backbone:lenet5}, \Cref{tab:backbone:mlp} and \Cref{tab:backbone:resnet12} show the backbones used to construct the Bayesian template network used in our NAS. These backbone structures are also the network structures we used to construct other BNN baselines. 

\begin{table}[h!]
	\caption{
	  Details of the LeNet5 NAS backbone}
	\label{tab:backbone:lenet5}
	\begin{center}
	\adjustbox{max width=1.\textwidth}{%
	\begin{tabular}{ccc}
	  \toprule
	  Layer Name
	  & Base channel counts 
	  & Stride
	  \\
	  \midrule
	  Conv0 
	  & 64 
	  & 2
	  \\
	  Conv1
	  & 64 
	  & 2
	  \\
	  Linear0 
	  & 128 
	  & -
	  \\
	  Linear1 
	  & 128
	  & -
	  \\
	  Linear2
	  & 10
	  & -
	  \\
	  \bottomrule
	  \end{tabular}
      }
	\end{center}
	\vskip -0.1in
      \end{table}

      \begin{table}[h!]
	\caption{
	  Details of the MLP NAS backbone}
	\label{tab:backbone:mlp}
	\begin{center}
	\adjustbox{max width=1.\textwidth}{%
	\begin{tabular}{ccc}
	  \toprule
	  Layer Name
	  & Base channel counts 
	  & Stride
	  \\
	  \midrule
	  Linear0 
	  & 32
	  & -
	  \\
	  Linear1
	  & 32
	  & -
	  \\
	  Linear2
	  & 32
	  & -
	  \\
	  Linear3
	  & 2
	  & -
	  \\
	  \bottomrule
	  \end{tabular}
      }
	\end{center}
	\vskip -0.1in
      \end{table}

      \begin{table}[h!]
	\caption{
	  Details of the ResNet-based NAS backbone}
	\label{tab:backbone:resnet12}
	\begin{center}
	\adjustbox{max width=1.\textwidth}{%
	\begin{tabular}{ccc}
	  \toprule
	  Layer Name
	  & Base channel counts 
	  & Stride
	  \\
	  \midrule
	  Block0\_Layer0 
	  & 32 
	  & 2 \\
	  Block0\_Layer1
	  & 32 
	  & 1 \\
	  Block0\_Layer2
	  & 32 
	  & 1
	  \\
	  Block1\_Layer0 
	  & 64
	  & 2 \\
	  Block1\_Layer1
	  & 64 
	  & 1 \\
	  Block1\_Layer2
	  & 64 
	  & 1 \\
	  Block2\_Layer0 
	  & 128
	  & 2 \\
	  Block2\_Layer1
	  & 128
	  & 1 \\
	  Block2\_Layer2
	  & 128
	  & 1 \\
	  Block3\_Layer0 
	  & 256
	  & 2 \\
	  Block3\_Layer1
	  & 256
	  & 1 \\
	  Block3\_Layer2
	  & 256
	  & 1 \\
	  \bottomrule
	  \end{tabular}
      }
	\end{center}
	\vskip -0.1in
    \end{table}

\section{NAS controller MLP}
\label{sec:appendix:controller_mlp}
As mentioned in \Cref{sec:method:nas}, the NAS controller builds based on trainable embedding $z$ that is then encoded using an MLP. The trainable free variable $z$ is a vector of size $256$  and we use a four-layer MLP with ReLU activations and $512$ hidden units for each single layer. The output of the MLP is fed to multiple linear layers $f_{l,k}$ where the dimension of the layer will match the number of the possible candidates.

\section{Noise annealing for the NAS controller}
\label{sec:appendix:noise}
As mentioned in \Cref{sec:method:nas}, the NAS controller outputs a series of post-softmax vectors and pick suitable architectural operations based on these probabilities. Empirically, we found that adding noise to the post-softmax probabilities is essential to avoid falling into local minima in the architecture search space. We used the following noise generation process:

\begin{equation}
	p = 
	\begin{cases}
	\lambda_n \times f_n(\mu=0, \sigma=1.0) 	&  m \leq M_{w} \\
	p_{softmax} + \lambda_n \times f_n(\mu=0, \sigma=1.0) \times(M_{total} - m) & {otherwise}

	\end{cases}
\end{equation}

We found $f_n$ can either be a Gaussian noise truncated to only contain the positive values or a log-normal distribution. In practice, we used the Gaussian noise with truncation. $\lambda_n$ controls how aggressive the noise generation is and we linearly decrease the amount of noise after $M_{w}$ number of epochs. $m$ is the number of epoch and $M$ is the total number of search epochs that is normally set to $100$ in our experiments.

\begin{figure}[!h]
	\centering
	\includegraphics[width=.8\linewidth]{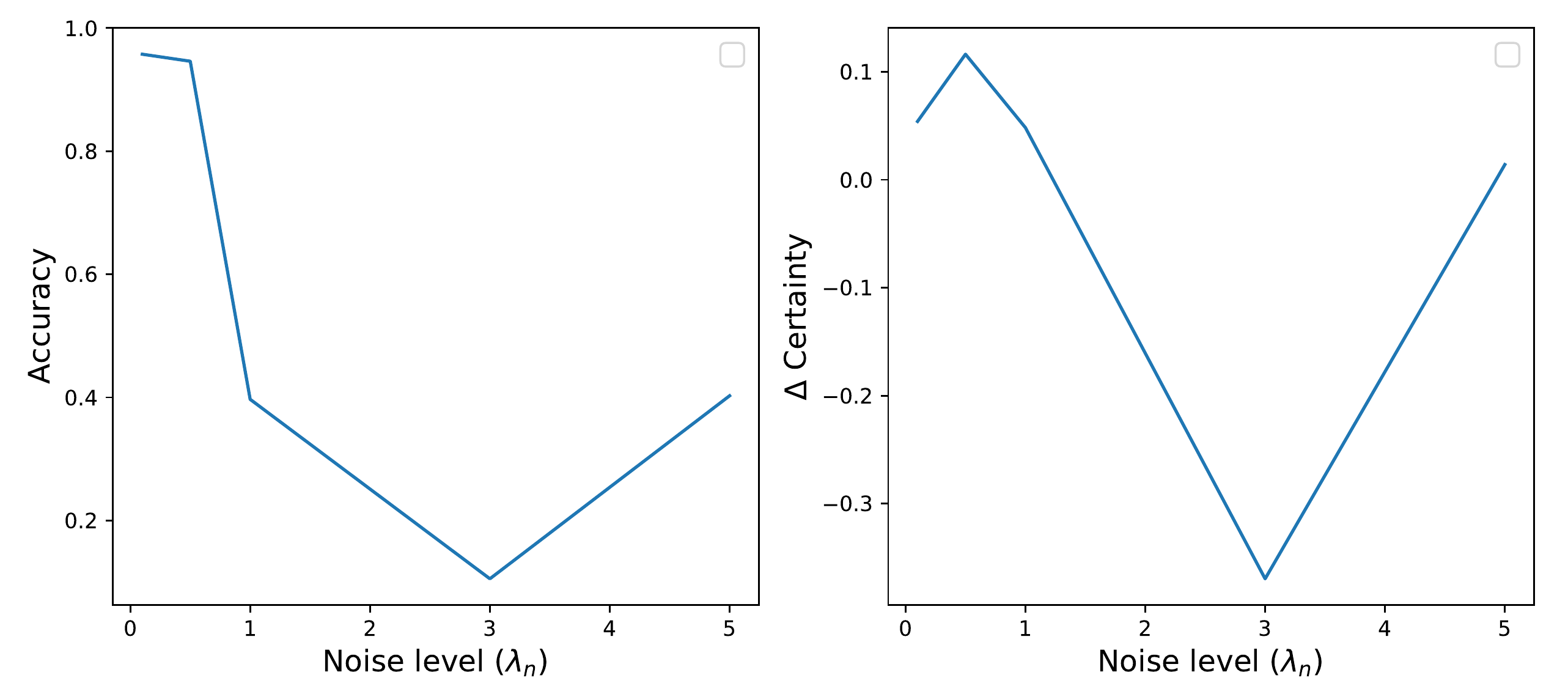}
	\caption{Accuracy and $\Delta$ Certainty tradeoffs with different noise levels ($\lambda$) used in the NAS controller on MNIST, with $N_{warmup} = 20$.  }
	\label{fig:nas:noise}
\end{figure}

\begin{figure}[!h]
	\centering
	\includegraphics[width=.8\linewidth]{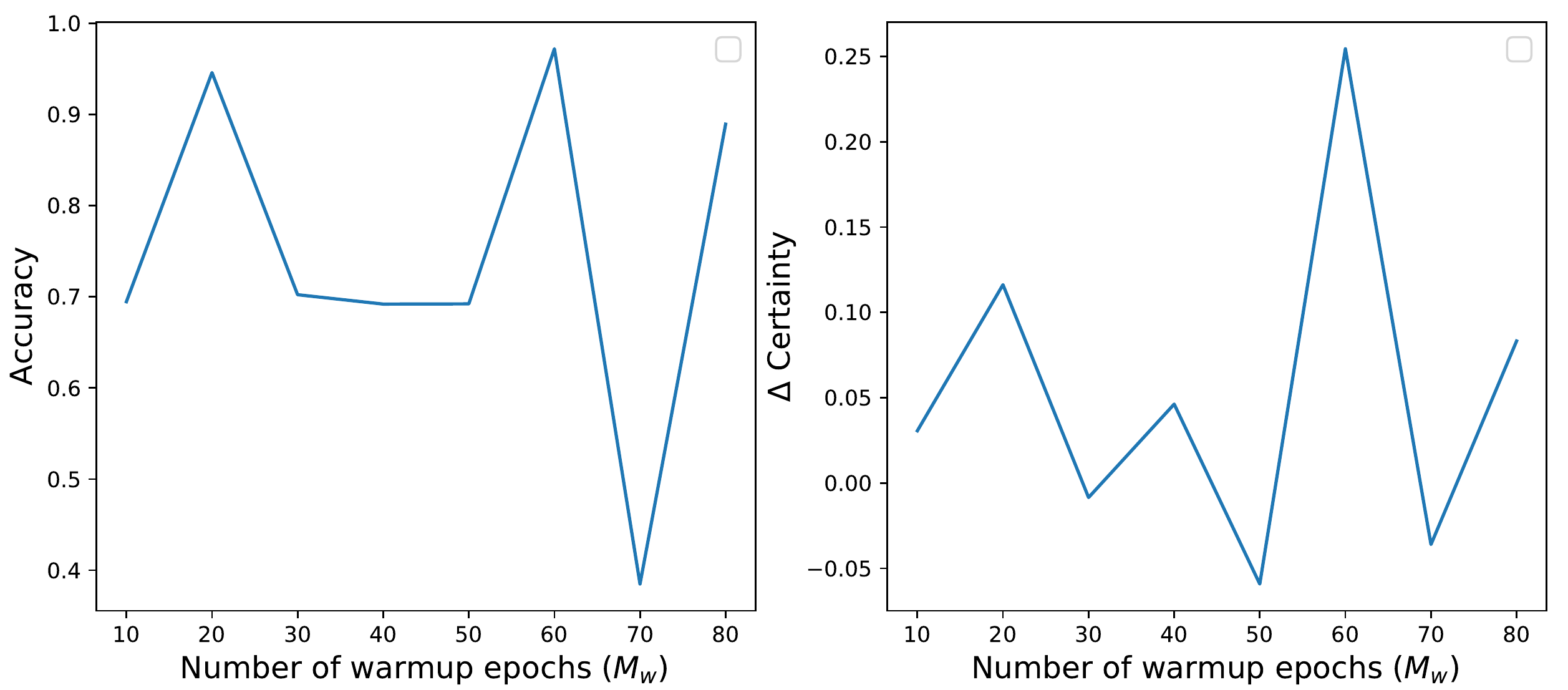}
	\caption{Accuracy and $\Delta$ Certainty tradeoffs with different number of warmup epochs ($M_{w}$) used in the NAS controller on MNIST, with $\lambda_n = 0.1$.  }
	\label{fig:nas:warmup}
\end{figure}

\Cref{fig:nas:noise} and \Cref{fig:nas:warmup} show our hyper-parameter studies on different $\lambda_n$ and $M_{w}$ values. We then picked $\lambda_n = 0.1, M_{w} = 20$ for the best trade-off between accuracy and uncertainty performance.
These two parameters together tunes the amount of exploration happens at the search stage. There should be an exploration for the NAS algorithm to avoid converge to local minima, but the exploration has to be limited so that finally the search process converges.

\section{VAE training setup}
\label{sec:appendix:vae}
To generate o.o.d data, we use a VAE to embed the training dataset and produce a reconstructed, noisy o.o.d dataset.

We train the VAE for 100 epochs with standard data augmentations including random affine transformations and Gaussian noising where it is applicable.
The VAE structure for Loan and Heart disease prediction contains a four-layer encoder, four-layer decoder architecture; each layer in this VAE contains 128 hidden units.
The VAE used for image datasets (MNIST and CIFAR) contains four convolutional layers for encoder and four convolutional layers for decoder.
\Cref{tab:vae} illustrates the model architecture of the VAE, and we use $n=32$ for MNIST and $n=64$ for CIFAR10.
The binary cross entropy loss is used for image datasets, and Mean-squared-error loss is used for other prediction tasks. We found that an Adam optimizer with a learning rate of $1e^{-4}$ is suitable for training the VAE on various datasets. 

\begin{table}[h!]
	\caption{
	VAE model on the MNIST and CIFAR10 dataset, we use $n=32$ for MNIST and $n=64$ for CIFAR10. The final layer will have 3 channels for CIFAR10 and 1 channel for MNIST.}
	\label{tab:vae}
	\begin{center}
	\adjustbox{max width=\textwidth}{%
	\begin{tabular}{cccc}
	 \toprule
	  Layer Name
	  & channel counts 
	  & Stride
	  & Kernel size
	  \\
	  \midrule
	  Encode\_Conv0 
	  & $n$
	  & 2
	  & 3
	  \\
	  Encode\_Conv1
	  & $n \times 2$
	  & 2
	  & 3
	  \\
	  Encode\_Conv2
	  & $n \times 4$
	  & 2
	  & 3
	  \\
	  Encode\_Conv3
	  & $n \times 4$
	  & 2
	  & 3
	  \\
	  Decoder\_Deconv0
	  & $n \times 4$
	  & 2
	  & 3
	  \\
	  Decoder\_Deconv1
	  & $n \times 4$
	  & 2
	  & 3
	  \\
	  Decoder\_Deconv2
	  & $n \times 2$
	  & 2
	  & 3
	  \\
	  Decoder\_Deconv3
	  & 3/1
	  & 2
	  & 3
	  \\
	  \bottomrule
	  \end{tabular}
      }
	\end{center}
\end{table}

\section{Bayesian inference on the n-last layer(s)}
\label{sec:appendix:inference}
\begin{figure}[!h]
	\centering
	\includegraphics[width=\linewidth]{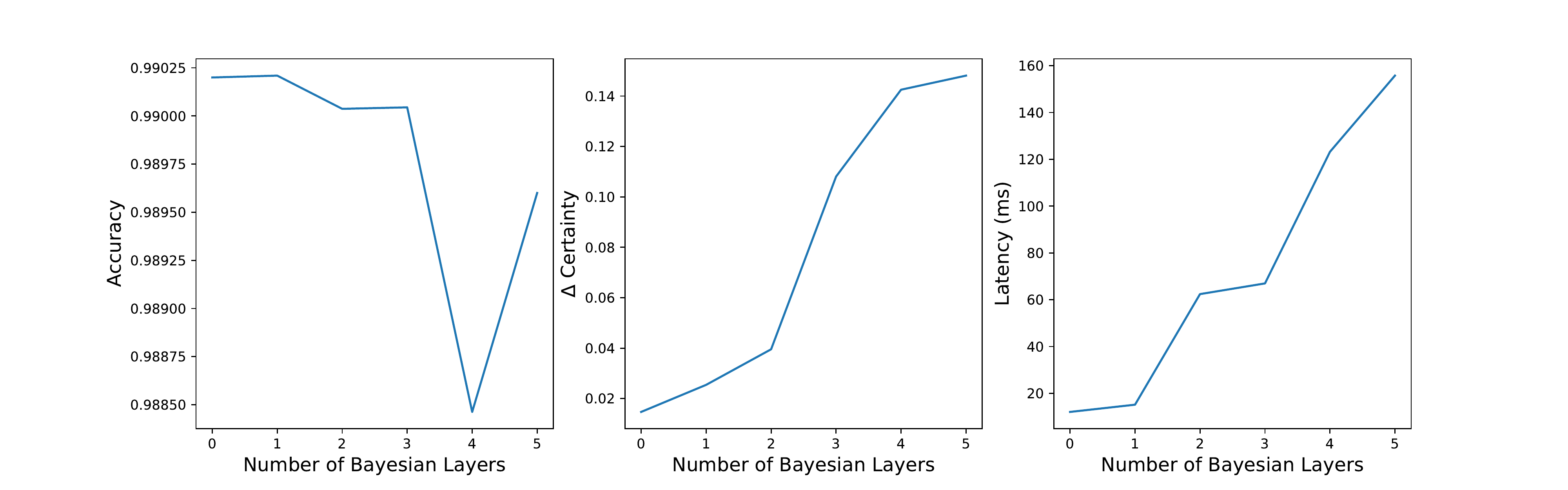}
	\caption{Accuracy $\Delta$ Certainty and Latency tradeoffs with different number of Bayesian layers on a Lenet5 based architecture with LRT.  }
	\label{fig:inference}
\end{figure}

A common design practice to reduce the runtime overhead of BNNs is to only use the later portion of the network to perform Bayesian inference.
While \Cref{sec:eval:efficiency} discussed the inference efficiency difference between the NAS searched models and hand-designed models. In \Cref{fig:inference} we illustrate the details of the hand-designed models.
This plot shows how having different numbers of Bayesian layers can affect accuracy, uncertainty measurements and latency.

\section{The effect of different learning rates}
\label{sec:appendix:lr}
The NAS algorithm contains a bi-level optimization as mentioned in our main paper. We use two Adam optimizers to update the Bayesian template networks and the NAS controller respectively. There then exist two learning rates, one for training the backbone network ($lr_t$) and the other for the NAS controller $lr_{arch}$.
As mentioned in \Cref{sec:exp:overview}, we pick the learning rates from $\{1e^{-2}, 1e^{-3}, 1e^{-4}\}$.
Our results in \Cref{tab:lr_ablation} show that a suitable learning rate combination for our NAS algorithm is $lr_t=1e^{-4}, lr_{arch} = 1e^{-3}$, we use this learning rate combination for all other datasets in the main paper.
\begin{table*}[ht!]
	\caption{
		Running the NAS algorithm with different learning rates for MNIST classification, results are averaged across three independent runs. $lr_t$ and $lr_{arch]}$ represent the learning rate for the Bayesian template network and the learning rate for the NAS controller respectively.
	}
	\label{tab:lr_ablation}
	\begin{center}
	\adjustbox{scale=.9}{%
	\begin{tabular}{c|ccc}
	\toprule
	\multirow{1}*{}   
	& \multicolumn{3}{c}{$lr_t$}\\
	\midrule
	$lr_{arch}$
	& $1e^{-2}$	& $1e^{-3}$	& $1e^{-4}$\\
	\midrule
	$1e^{-2}$	
	& $0.3566 \pm 0.0061$ 
	& $0.9689 \pm 0.0008$
	& $0.9901 \pm 0.0005$ 
	\\
	$1e^{-3}$ 
	& $0.7728 \pm 0.0391$ 
	& $0.9883 \pm 0.0018$
	& $\mathbf{0.9910 \pm 0.0011}$\\
	$1e^{-4}$ 
	& $0.8566 \pm 0.0021$ 
	& $0.9733 \pm 0.0002$
	& $0.9823 \pm 0.0019$\\
	\bottomrule
	\end{tabular}
	}
	\end{center}
	\vskip -0.1in
    \end{table*}

\bibliographystyle{unsrtnat}
\bibliography{references}  






\end{document}